\documentclass[sigconf]{acmart}

\AtBeginDocument{%
  }

\copyrightyear{2023}
\acmYear{2023}
\setcopyright{acmlicensed}
\acmConference[MM '23]{Proceedings of the 31st ACM International Conference on Multimedia}{October 29-November 3, 2023}{Ottawa, ON, Canada}
\acmBooktitle{Proceedings of the 31st ACM International Conference on Multimedia (MM '23), October 29-November 3, 2023, Ottawa, ON, Canada}
\acmPrice{15.00}
\acmDOI{10.1145/3581783.3612451}
\acmISBN{979-8-4007-0108-5/23/10}

\acmSubmissionID{3280}

\usepackage{xspace}
\usepackage{subcaption}
\usepackage{siunitx}
\usepackage{cleveref}
\usepackage{wrapfig}
\usepackage{stmaryrd}
\usepackage{amsfonts}
\usepackage{bm}
\usepackage{amsmath}
\usepackage{multirow}
\usepackage{graphicx}
\usepackage{hyperref}

\begin{document}

\title{Painterly Image Harmonization using Diffusion Model}

\author{Lingxiao Lu}
\affiliation{%
    \institution{MoE Key Lab of Artificial Intelligence, Shanghai Jiao Tong University}
    \country{China}
}
\email{lulingxiao@sjtu.edu.cn}

\author{Jiangtong Li}
\affiliation{%
    \institution{MoE Key Lab of Artificial Intelligence, Shanghai Jiao Tong University}
    \country{China}
}
\email{keep_moving-lee@sjtu.edu.cn}

\author{Junyan Cao}
\affiliation{%
    \institution{MoE Key Lab of Artificial Intelligence, Shanghai Jiao Tong University}
    \country{China}
}
\email{joy_c1@sjtu.edu.cn}

\author{Li Niu}
\authornote{Corresponding authors}
\affiliation{%
    \institution{MoE Key Lab of Artificial Intelligence, Shanghai Jiao Tong University}
    \country{China}
}
\email{ustcnewly@sjtu.edu.cn}

\author{Liqing Zhang}
\authornotemark[1]
\affiliation{%
    \institution{MoE Key Lab of Artificial Intelligence, Shanghai Jiao Tong University}
    \country{China}
}
\email{zhang-lq@cs.sjtu.edu.cn}

\begin{abstract}
Painterly image harmonization aims to insert photographic objects into paintings and obtain artistically coherent composite images. Previous methods for this task mainly rely on inference optimization or generative adversarial network, but they are either very time-consuming or struggling at fine control of the foreground objects (\emph{e.g.}, texture and content details). To address these issues, we propose a novel Painterly Harmonization stable Diffusion model (PHDiffusion), which includes a lightweight adaptive encoder and a Dual Encoder Fusion (DEF) module. Specifically, the adaptive encoder and the DEF module first stylize foreground features within each encoder. Then, the stylized foreground features from both encoders are combined to guide the harmonization process. During training, besides the noise loss in diffusion model, we additionally employ content loss and two style losses, \emph{i.e.}, AdaIN style loss and contrastive style loss, aiming to balance the trade-off between style migration and content preservation. Compared with the state-of-the-art models from related fields, our PHDiffusion can stylize the foreground more sufficiently and simultaneously retain finer content. Our code and model are available at \href{https://github.com/bcmi/PHDiffusion-Painterly-Image-Harmonization}{https://github.com/bcmi/PHDiffusion-Painterly-Image-Harmonization}.
\end{abstract}



\maketitle

\section{Introduction}

The goal of painterly image harmonization is to integrate photographic objects into background paintings and achieve visual coherence. 
While standard image harmonization~\cite{cong2020dovenet,cong2022high} focuses on adapting low-level statistics (\emph{e.g.}, color, brightness), painterly image harmonization~\cite{peng2019element, zhang2020deep, cao2022painterly} is more challenging as it requires transferring high-level styles in addition to low-level statistics.

Existing works for this task can be roughly divided into two categories:  optimization-based~\cite{luan2018deep, zhang2020deep} and feed-forward~\cite{peng2019element, cao2022painterly, yan2022style} approaches. 
For the optimization-based approaches~\cite{luan2018deep, zhang2020deep}, they optimize over the composite image by minimizing the designed losses, which makes them very time-consuming and unsuitable for real-time applications.
Besides, the feed-forward~\cite{peng2019element, cao2022painterly, yan2022style} methods mainly rely on Generative Adversarial Network (GAN)~\cite{goodfellow2020generative} and the trained model can directly generate the harmonized images. 
However, one limitation of GAN-based approaches is limited control over complex foregrounds~\cite{shoshan2021gan}, resulting in unsatisfactory harmonized foregrounds (\emph{e.g.}, loss of content and style details).

\begin{figure}[t]
\begin{minipage}{0.23\linewidth}
\centerline{\includegraphics[width=1\textwidth]{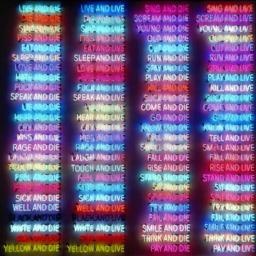}}
\centerline{Background}
\end{minipage}
\hspace{0.1mm}
\begin{minipage}{0.23\linewidth}
\centerline{\includegraphics[width=1\textwidth]{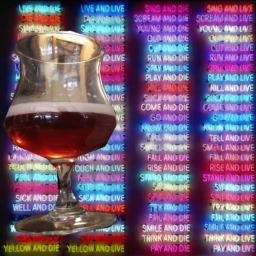}}
\centerline{Cut-and-Paste}
\end{minipage}
\hspace{0.1mm}
\begin{minipage}{0.23\linewidth}
\centerline{\includegraphics[width=1\textwidth]{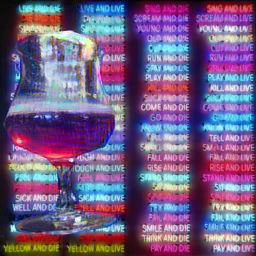}}
\centerline{PHDNet}
\end{minipage}
\hspace{0.1mm}
\begin{minipage}{0.23\linewidth}
\centerline{\includegraphics[width=1\textwidth]{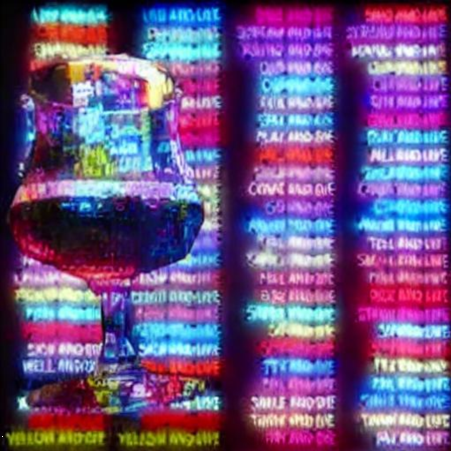}}
\centerline{Ours}
\end{minipage}
\caption{Painterly image harmonization aims to harmonize the inserted photographic foreground according to the background painting. From left to right, we present the background image, the composite image via cut-and-paste, the harmonized results of PHDNet~\cite{cao2022painterly} and our method. }

\end{figure}

In recent years, diffusion models~\cite{ho2020denoising} have demonstrated comparable or better performance compared with state-of-the-art image generation models, by formulating image generation as sequential stochastic transitions from a simple distribution to data distribution.
Diffusion methods can be divided into unconditional diffusion methods~\cite{ho2020denoising, song2020denoising} and conditional diffusion methods~\cite{rombach2022high, mou2023t2i}. 
The unconditional diffusion methods aim to generate realistic images by modeling the distribution of natural images without conditioning on any specific input.
Whereas, the conditional diffusion methods aim to generate images with the guidance of conditional information~(\emph{e.g.}, text, semantic mask, \emph{etc.}).
Among them, Stable Diffusion (SD)~\cite{rombach2022high} is one of the most popular models, which successfully integrates the text CLIP~\cite{radford2021learning} into latent diffusion.
Further, some more recent models (\emph{e.g.}, T2I-adapter~\cite{mou2023t2i}, ControlNet~\cite{zhang2023adding}) freeze the SD model and introduce trainable adapters to encode different types of conditions into SD through multi-step guidance.
Therefore, conditional diffusion models offer a promising and flexible approach to improve painterly image harmonization by enabling multi-step guidance for the photographic foreground.

Several existing works have introduced diffusion methods into similar tasks like cross-domain image composition~\cite{hachnochi2023cross} and image editing~\cite{meng2021sdedit}. 
For example, CDC~\cite{hachnochi2023cross} proposed an inference-time conditioning method that uses high-frequency details from the background and low-frequency style from the foreground object for image composition. 
However, CDC~\cite{hachnochi2023cross} assumes that high-frequency (\emph{resp.}, low-frequency) feature in the image represents style (\emph{resp.}, content) information, which does not always hold.
Another work SDEdit~\cite{meng2021sdedit} synthesizes images by first adding noise to the input image and then iteratively denoising through a stochastic differential equation. 
However, this approach lacks proper and sufficient guidance during the denoising process, leading to the final image lacking sufficient styles and contents.

In this paper, we introduce Painterly Harmonization stable Diffusion model (PHDiffusion), which exploits two extra modules based on the Stable Diffusion (SD) model.
Inspired by the conditional diffusion model~\cite{mou2023t2i}, we equip SD with a lightweight adaptive encoder, which aims to extract the required condition information (\emph{i.e.}, background style, image content) from the composite image. As part of denoising U-Net, the denoising encoder in SD takes composite image as input. The adaptive encoder takes in the concatenation of composite image and foreground mask, producing residuals added to the feature maps in the denoising encoder. 
Based on the denoising encoder and the adaptive encoder, we introduce a Dual Encoder Fusion (DEF) module to fuse the information from two encoders.
Specifically, given the image features extracted by two encoders, our DEF module incorporates the background style into foreground content and generates the stylized foreground features.
Then, the stylized foreground features from two encoders are combined to provide multi-step guidance in the denoising steps.

To utilize the rich prior knowledge in pretrained SD and relieve the training burden, following~\cite{mou2023t2i}, we freeze the model parameters of SD, and only update the adaptive encoder and DEF module during training. 
The standard noise loss used in diffusion models~\cite{rombach2022high} could maintain the image content, but cannot migrate background style to the foreground. Therefore, we further introduce two additional style losses, \emph{i.e.}, AdaIN loss and contrastive style loss, to balance the style and content for foreground object. 
The AdaIN loss~\cite{huang2017arbitrary} aligns the multi-scale statistics (\emph{e.g.}, mean, variance) of the foreground object with the background painting, while the contrastive style loss~\cite{chen2021artistic} aims to push the foreground style towards background style.
In addition, we also incorporate a content loss to address the issue of excessive content preservation  by using noise loss alone.
With noise loss, style losses, and content loss, our PHDiffusion is able to comprehend the background style and preserve the foreground content. 
During testing, our PHDiffusion could be directly used to produce harmonized image, preventing additional time-consuming inference optimization~\cite{kwon2022diffusion,hachnochi2023cross}.

To verify the effectiveness of our PHDiffusion, we compare our methods with the state-of-the-art methods, and conduct experiments on the benchmark datasets COCO~\cite{lin2014microsoft} and WikiArt~\cite{nicholwikiart}.
\textbf{The experimental results show that our PHDiffusion can achieve certain visually pleasant results that previous methods cannot achieve, especially when the background has dense textures or abstract style.} 
Our contributions can be summarized as follows: 1) We are the first work focusing on painterly image harmonization using diffusion model. 2) We propose a Painterly Harmonization stable Diffusion model (PHDiffusion) by using dual encoder fusion to provide effective guidance and reasonable loss designs to achieve sufficient stylization. 3) The experimental results show that our PHDiffusion strikes a good balance between adapting styles and maintaining structures. 

\section{Related Work}

\subsection{Image Harmonization}
As a subtask of image composition~\cite{niu2021making}, image harmonization aims to adjust the color and illumination statistics of foreground to be compatible with background in a composite image. 
In recent years, deep learning methods~\cite{cong2021bargainnet,zhu2015learning,chen2019toward,guo2021intrinsic,xing2022composite} play an important role in this field. 
Especially after the first large-scale image harmonization dataset iHarmony4~\cite{cong2020dovenet} was released, supervised image harmonization~\cite{hao2020image, cun2020improving, ling2021region, hang2022scs,guo2021image,cong2021deep} methods have received more and more attention. 
For example, DoveNet~\cite{cong2020dovenet} approached image harmonization as a domain translation task. 
Hao~\emph{et al.}~\cite{hao2020image} utilized attention block to calculate non-local information for foreground adjustment. 
SSAM~\cite{cun2020improving} focused on relation between the spliced region and non-spliced region by exploiting a dual path attention model to fuse them together. 
CDTNet~\cite{cong2022high} combined pixel-to-pixel transformation and RGB-to-RGB transformation for high-resolution image harmonization. 
Recently, DCCF~\cite{xue2022dccf} and \(S^2CRNet\)~\cite{liang2022spatial} are applied in this field for high resolution image harmonization.
Note that the abovementioned methods require ground-truth image to supervise, which is not suitable for our task.

\subsection{Painterly Image Harmonization}
As a similar task to image harmonization, painterly image harmonization aims to blend a photographic foreground into an artistic background painting, resulting in a visually coherent painting. 
Compared with image harmonization, painterly image harmonization is more challenging as it needs to adapt high-level styles beyond low-level statistics.
Deep Painterly Harmonization~\cite{luan2018deep} introduced a two-pass algorithm to ensure both spatial and inter-scale statistical consistency. Meanwhile, Deep Image Blending~\cite{zhang2020deep} utilized a two-stage blending algorithm and proposed a Poisson blending loss to guide blending together with content and style loss. 
However, both Deep Painterly Harmonization~\cite{luan2018deep} and Deep Image Blending~\cite{zhang2020deep} are optimization-based method, which optimizes the input image during inference, making them unusable from real-time harmonization.
On the other hand, E2STN~\cite{peng2019element},  PHDNet~\cite{cao2022painterly}, and Yan~\emph{et al.}~\cite{yan2022style} exploited the feed-forward scheme by first training the generator and then directly producing the harmonized image during inference.
Specifically, E2STN~\cite{peng2019element} took advantages of both global and local discriminators to harmonize the embedded element with the background image. 
PHDNet\cite{cao2022painterly} exploited spatial and frequency domains to capture different types of background type, and then adjusted the foreground in both domains. 
Yan~\emph{et al.}~\cite{yan2022style} integrated  GP-GAN~\cite{wu2019gp}, WCT~\cite{li2017universal}, and StyleTr$^2$~\cite{deng2022stytr2} together to fuse the global and local information together.
Note that, these feed-forward approaches~\cite{peng2019element,cao2022painterly,yan2022style} are mainly based on adversarial learning by playing a minimax game between generator and discriminator, which have limited control over complex photographic foreground~\cite{shoshan2021gan} and have difficulty in leveraging prior knowledge across different image domains~\cite{goodfellow2020generative}. 
Different from them, our method is built upon the diffusion model, with stylized foreground features as guidance and a combo of style losses to produce the harmonized result. 

\subsection{Artistic Style Transfer}
Artistic style transfer aims to stylize a content image given a style image. 
Previous optimization-based methods~\cite{gatys2016image,kolkin2019style,du2020much} optimize the content image to match its style with the style image. 
In contrast, feed-forward~\cite{johnson2016perceptual,huang2017arbitrary,liu2021adaattn, park2019arbitrary, deng2022stytr2,li2016precomputed} methods generate stylized images by training a generator to combine the content image and the style image.
For example, style-relevant statistics~\cite{li2017demystifying,zhang2022exact} (\emph{e.g.}, mean and standard deviation of feature map) between the style image and fused image should be similar, and content-relevant information~\cite{donahue2014decaf} (\emph{e.g.}, the categories of objects) within the fused image should also be kept from the content image. 
Moreover, to enhance the visual quality in artistic style transfer, contrastive learning is also introduced~\cite{zhang2022domain,chen2021artistic} to capture sufficient style information. 
Artistic style transfer methods stylize the entire content image, while painterly image harmonization focuses on the inserted object in the background painting.

\subsection{Diffusion Models}
Recently, diffusion models have shown remarkable performance in image generation~\cite{ho2020denoising, song2020denoising, liu2022pseudo}, text-to-image generation~\cite{nichol2021glide,rombach2022high}, image translation~\cite{kwon2022diffusion}, image inpainting~\cite{lugmayr2022repaint,saharia2022palette}, and image editing~\cite{meng2021sdedit, zhang2022inversion, hachnochi2023cross, jeong2023training}.
Image editing and cross-domain image composition are the most relevant fields to our painterly image harmonization. Hence, we focus on these two fields with diffusion models in this section.
Specifically, SDEdit~\cite{meng2021sdedit} employed the image synthesis approach that commences with the addition of noise to the input image, followed by iterative denoising process with stochastic differential equation. 
CDC~\cite{hachnochi2023cross} proposed to harmonize the image in frequency domain by exploiting high-frequency details from the background and low-frequency style from the foreground object. 
Besides, there are a few diffusion models designed for artistic style transfer. 
To name a few, DiffStyle~\cite{jeong2023training} disentangled representations for content and style, and fused them in h-space, which lies in the bottleneck of U-Net. 
InST~\cite{zhang2022inversion} was motivated by the belief that an unique artwork can not be directly explained by words, so it designed an encoding module that maps style image into text domain through a CLIP image encoder.

However, previous diffusion-based methods can not provide powerful style guidance and hold adequate content in painterly image harmonization. 
In this work, we endow stable diffusion with stylized feature guidance and well-designed losses for adjusting sufficient styles and maintaining content details, leading to better harmonization performance.

\section{Method}
\label{sec:method}

\begin{figure*}[t]
  \centering
  \includegraphics[width=0.86\linewidth]{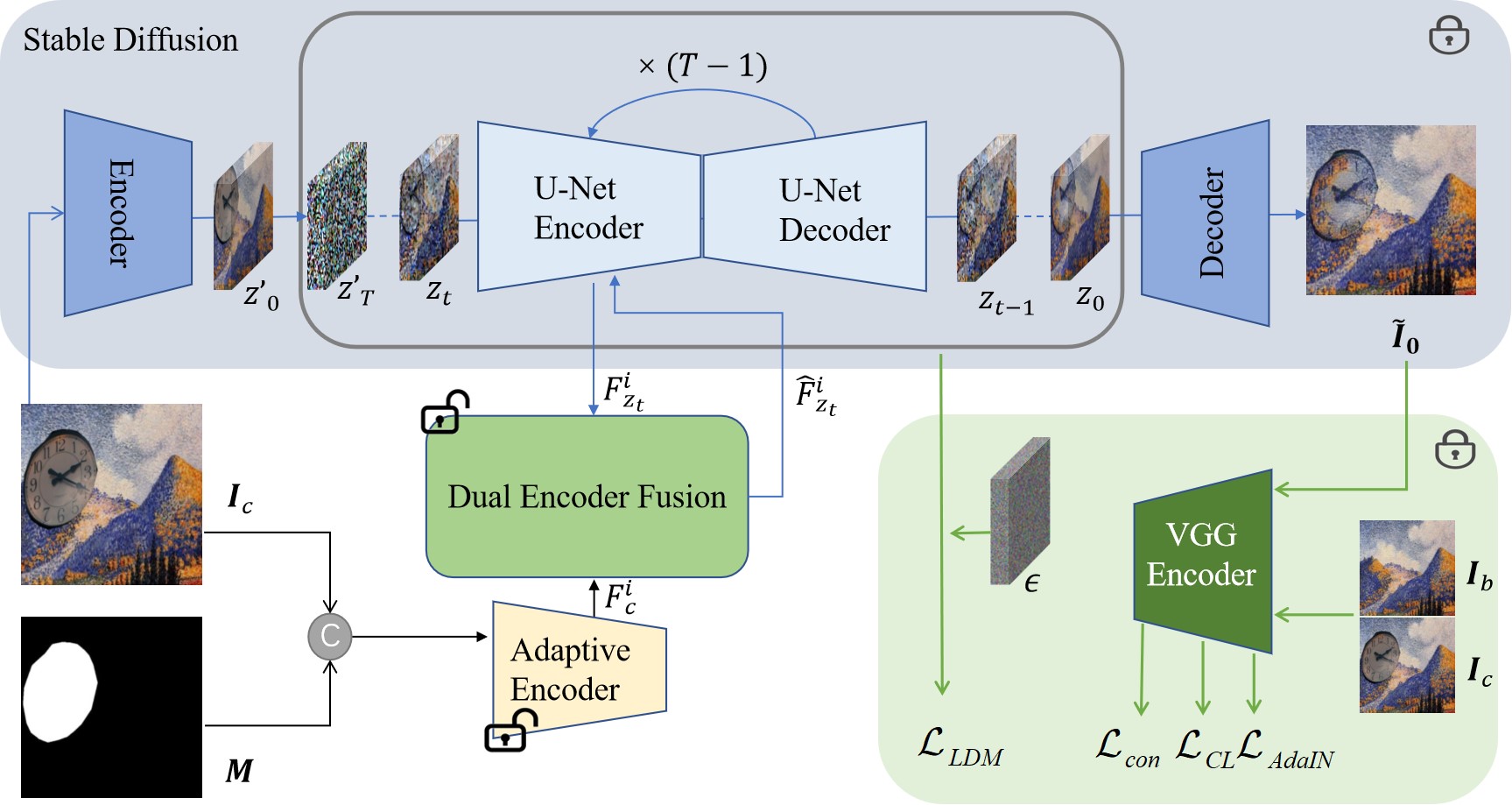}
  \caption{The architecture of our PHDiffusion. Given an composite image $\bm{I}_c$ and its foreground mask $\bm{M}$, $\bm{I}_c$ is sent to a pretrained Stable Diffusion~\cite{cong2022high} model for painterly image harmonization. The input $\bm{I}_c$ is first encoded to the latent space \(z'_0=\mathcal{E}(\bm{I}_c)\), which is followed by the forward process of diffusion to deduce \(z'_t; t=1,2,...T\) with noise~$\epsilon$. During inference, \(z'_T\) is used as the initial input \(z_T\) for the backward process to predict \(z_t; t=T-1,T-2,...0\) through the denoising U-Net. Finally, the harmonized image $\bm{\tilde{I}}_{0}$ is generated through the decoder by $\bm{\tilde{I}}_{0}=\mathcal{D}(z_0)$.
  In the meanwhile, the composite image $\bm{I}_c$ concatenated with foreground mask $\bm{M}$ is sent to the adaptive encoder, which is followed by the Dual Encoder Fusion to provide guidance to the denoising process in U-Net. 
  The denoising process is supervised by the noise loss~$\mathcal{L}_{LDM}$. 
  Besides, we exploit two style losses ($\mathcal{L}_{AdaIN}$ and $\mathcal{L}_{CL}$) for foreground stylization and content loss ($\mathcal{L}_{con}$) for content preservation.}
  \label{fig:main_network}
  \Description{Architecture of network.}
\end{figure*}

The overall framework of our PHDiffusion is depicted in \Cref{fig:main_network}, which consists of a Stable Diffusion (SD) model, an adaptive encoder, and a Dual Encoder Fusion (DEF) module. 
Given a composite image~$\bm{I}_c$ and the corresponding foreground mask~$\bm{M}$, we first exploit the adaptive encoder to extract the multi-scale composite feature maps~$\bm{F}_c^i, \  i \in \{1,2,3,4\}$.
In order to guide the generation of SD, we fuse the composite feature map~$\bm{F}_c^i$ with the corresponding denoising feature map~$\bm{F}_{z_t}^i$ from the U-Net encoder of SD according to their resolutions through the DEF module to generate new denoising feature map~$\bm{\hat{F}}_{z_t}^i$.
Note that, we freeze the SD, and supervise the adaptive encoder and the DEF module using noise loss, content loss, and two style losses (AdaIN loss and contrastive style loss).  

Next, we will first briefly review SD model, introduce the adaptive encoder, and then elaborate on our DEF Module. 
Finally, we will introduce the objectives for training adaptive encoder and DEF.
As for the notation in the remainder of this section, \(z'_0\) is the initial latent feature extracted by encoder from image \(\bm{I}\). \(z'_t; t=1,2,...T\) represents the latent feature that is deduced from \(z'_0\) in the forward process of diffusion \(q(z'_t|z'_0)\).
While \(z_t; \ t \in \{T-1,...0\}\) is predicted in the backward denoising process. \(\bm{\tilde{I}}_{0}\) is the decoded harmonized image from \(z_0\). \(\hat{\bm{M}}\) is the mask that is down-sampled to size of \(z'_0\).

\subsection{Preliminaries}

Our method is built upon the Stable Diffusion (SD)~\cite{rombach2022high} model, where SD is a latent diffusion model pretrained in two stages comprised of an auto-encoder and a denoising U-Net. 
In the first stage, the SD model trains the auto-encoder, in which the encoder \(\mathcal{E}\) first encodes images \(\bm{I}\) into latent space \(z'_0=\mathcal{E}(\bm{I})\) and then the decoder \(\mathcal{D}\) reconstructs them into original images \(\bm{\hat{I}}=\mathcal{D}(z'_0)\). 
In the second stage, the auto-encoder is frozen and the SD constructs the denoising U-Net~\(\epsilon_{\theta}\)~\cite{ho2020denoising} by first adding $T$-step noise to latent space feature~\(z'_0\) to generate \(z'_t; t=1,2,...T\), and then training the denoising U-Net with latent denoising loss, which is formulated as
\begin{equation} \label{eqn:prelim}
\mathcal{L}_{LDM}:=\mathbb{E}_{z'_0, y, \epsilon \sim \mathcal{N}(0,1), t}\left[\left\|\epsilon-\epsilon_{\theta_1}\left(z'_{t}, t,\tau_{\theta_2}(y) \right)\right\|_{2}^{2}\right],
\end{equation}
where $\epsilon$ is the noise that is added in latent space feature~\(z'_0\) in each noising step, $\epsilon_{\theta_1}$ is the denoising U-Net that predicts the noise~$\epsilon$ in current step $t$, \(y\) stands for extra condition (\emph{e.g.}, text, mask, \emph{etc.}), and \(\tau_{\theta_2}\) is a domain specific encoder that projects \(y\) to  intermediate representation. 
In this work, we add condition information, \emph{i.e.}, the composite image with foreground mask, using an adaptive encoder similar to~\cite{mou2023t2i}.

During inference, noise is first added to \(z'_0\) to generate \(z'_T\), and then \(z'_T\) is used as \(z_T\), the initial input for \(\epsilon_{\theta_1}\). \(\epsilon_{\theta_1}\) is then iteratively used to estimate the noise at each denoising step \(t\), thus the latent map \(z_T\) is gradually refined and ultimately becomes clean latent feature \(z_0\).
Finally, the clean latent feature \(z_0\) is fed into the decoder \(\mathcal{D}\) to generate the image. 
For more details about the training and inference of Stable Diffusion, please refer to~\cite{rombach2022high}. 

\subsection{Adaptive Encoder}

As introduced before, the adaptive encoder accounts for encoding additional condition and providing multi-step guidance in the denoising steps.
Previous implementations of the adaptive encoder~\cite{mou2023t2i} focus more on rough structures (\emph{e.g.}, sketch, pose, semantic mask), and exploit the text condition~\cite{rombach2022high} to indicate extra demands (\emph{e.g.}, styles or environments).
Different from previous works, we discard the text CLIP model and adopt the lightweight adaptive encoder~\cite{mou2023t2i} to encode the concatenation of composite image and foreground mask, simultaneously preserving content details and extracting background styles.
In detail, the architecture of adaptive encoder 
comprises of four feature extraction blocks and three DownSample (DS for short) blocks. The input with resolution \(512 \times 512\) is first downsampled to \(64 \times 64\) (named as $\bm{F}_c^0$) through pixel unshuffle~\cite{shi2016real}.
By combining one convolutional layer and two residual blocks as an extraction module (EM for short), the generation of $\bm{F}_c^i, \  i \in \{1,2,3,4\}$ can be formulated as:
\begin{equation}
    \begin{aligned}
        \bm{F}_c^1 &= \text{EM}_1(\bm{F}_c^0), \quad\quad\ \ \ \  \bm{F}_c^2 = \text{EM}_2(\text{DS}(\bm{F}_c^1)), \\
        \bm{F}_c^3 &= \text{EM}_3(\text{DS}(\bm{F}_c^2)), \quad \bm{F}_c^4 = \text{EM}_4(\text{DS}(\bm{F}_c^3)), \\
    \end{aligned}
\end{equation}
where the resolutions of $\bm{F}_c^1$, $\bm{F}_c^2$, $\bm{F}_c^3$, and $\bm{F}_c^4$ are \(64 \times 64\), \(32 \times 32\), \(16 \times 16\), and \(8 \times 8\), respectively. Similar structures also exist in U-Net encoder to generate $\bm{F}_{z_t}^i$ with the same resolution as $\bm{F}_c^i$, $\  i \in \{1,2,3,4\}$.

\subsection{Dual Encoder Fusion Module}
In the section, we introduce our Dual Encoder Fusion~(DEF) module to preserve content of foreground object and extract reasonable style from background painting. 
As shown in \Cref{fig:main_network}, the adaptive encoder first takes composite image \(\bm{I}_c\) and foreground mask \({\bm{M}}\) as input, then generates the composite feature maps \(\bm{F}_c^i, \  i \in \{1,2,3,4\}\) with different resolutions. 
Each of the composite feature maps~\(\bm{F}_c^i\) is then fused with the corresponding denoising feature maps~\(\bm{F}_{z_t}^i\).
However, we find that if we directly add or concatenate the feature maps from these two encoders, the final generation result cannot stylize the foreground well, leading to notable style discrepancy between foreground and background. 

Considering that CNN can only expand the receptive field of the foreground features within a certain range and struggles to capture long-range dependency~\cite{dosovitskiy2020image}, we propose to endow the foreground features with global receptive field to some extent.
To balance the global-local receptive field of the foreground features, we design different fusion strategies for the features with different resolutions. 
For shallow features with high resolutions, where \(i\in \{1,2\}\), composite feature maps \(\bm{F}_c^i\) are simply added to \(\bm{F}_{z_t}^i\) to maintain the local structures. 
While for deeper features with low resolutions, where \(i\in\{3,4\}\), \(\bm{F}_c^i\) and \(\bm{F}_{z_t}^i\) are fused through our DEF module to capture the global styles. 
The above process can be formulated as
\begin{equation}
\hat{\bm{F}}_{z_t}^i=
    \begin{cases}
    \bm{F}_c^i+\bm{F}_{z_t}^i,& i=1,2, \\
    \text{DEF}(\bm{F}_c^i, \bm{F}_{z_t}^i),& i=3,4 .
    \end{cases}
\end{equation}

\begin{figure}[t]
  \centering
  \includegraphics[width=\linewidth]{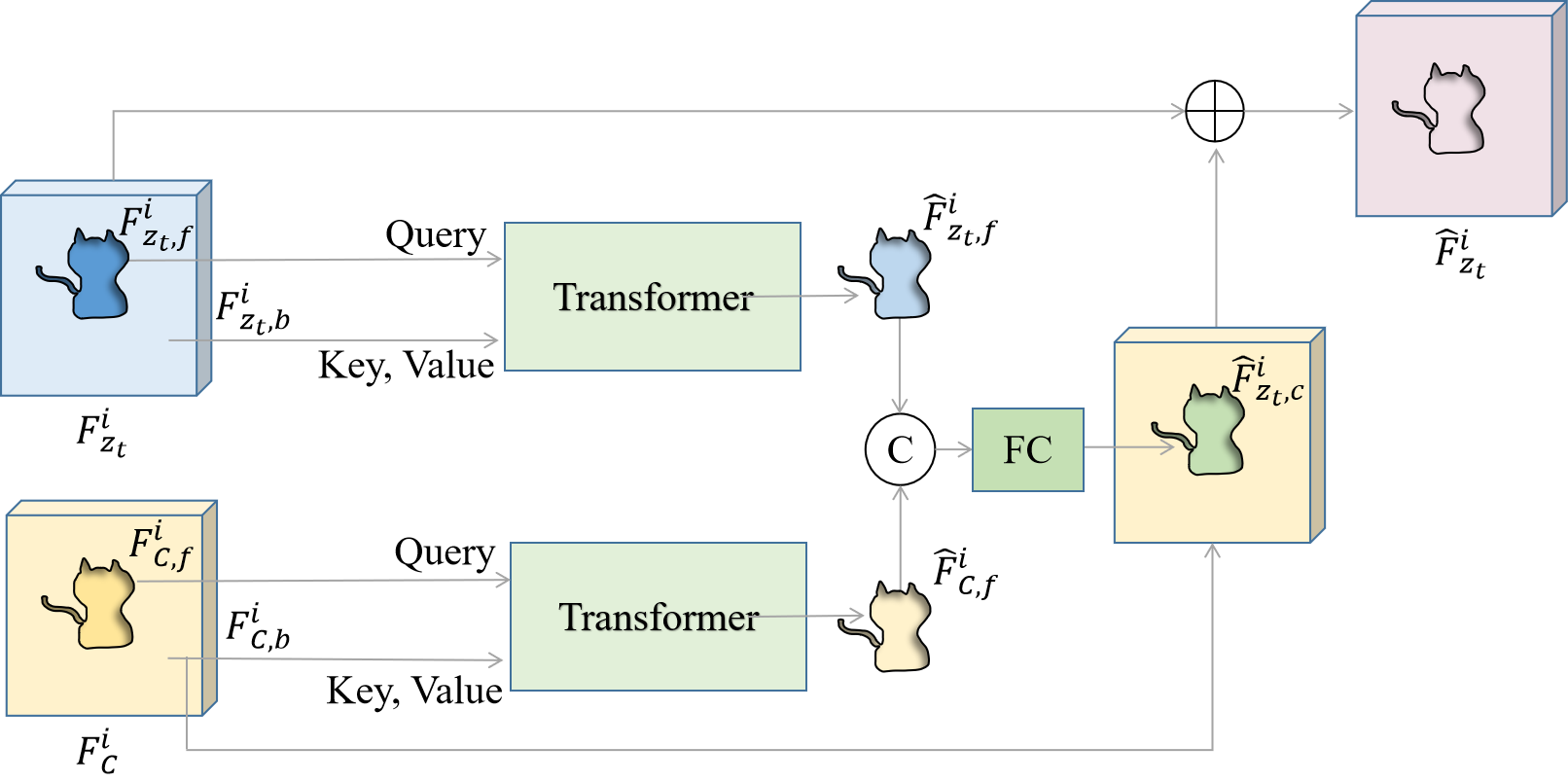}
  \caption{The architecture of Dual Encoder Fusion (DEF) module. Given the composite feature map $\bm{F}_{c}^i$ (\emph{resp.}, the denoising feature map $\bm{F}_{z_t}^i$), we stylize the foreground of composite feature map (\emph{resp.}, the denoising feature map) by regarding the foreground feature map $\bm{F}_{c, f}^i$ (\emph{resp.}, $\bm{F}_{z_t, f}^i$) as query and the background feature map $\bm{F}_{c, b}^i$ (\emph{resp.}, $\bm{F}_{z_t, b}^i$) as key/value 
 through a transformer layer to obtain the stylized foreground feature map $\bm{\hat{F}}_{c, f}^i$ (\emph{resp.}, $\bm{\hat{F}}_{z_t, f}^i$). Then, $\bm{\hat{F}}_{c, f}^i$ and $\bm{\hat{F}}_{z_t, f}^i$ are fused through concatenation and a fully-connected layer, which is then combined with the composite background feature map $\bm{F}_{c, b}^i$ and denoising feature map $\bm{F}_{z_t}^i$ to further guide the denoising process.
  }
  \label{fig:DEF}
\end{figure}

The structure of our DEF module is illustrated in \Cref{fig:DEF}, which consists of stylized feature extraction and stylized feature fusion.

\subsubsection{Stylized Feature Extraction}
Before we fuse composite feature maps~\(\bm{F}_c^i\) and denoising feature maps~\(\bm{F}_{z_t}^i\), we first need to expand the receptive field of the foreground features and extract the desired style from background.
We utilize the foreground features to search for relevant background styles through a transformer layer~\cite{vaswani2017attention}. 
In detail, by taking composite feature map~\(\bm{F}_c^i\) as example, we first extract its foreground features \(\bm{F}_{c,f}^i\) and background features \(\bm{F}_{c,b}^i\) by masking and flattening, which can be formulated as
\begin{equation}
    \begin{aligned}
        \bm{F}_{c, f}^i&=  \text{Flatten}(\bm{F}_c^i \circ \hat{\bm{M}}), \\
        \bm{F}_{c, b}^i&=  \text{Flatten}(\bm{F}_c^i \circ (1-\hat{\bm{M}})), 
    \end{aligned}
\end{equation}
where \(\hat{\bm{M}}\) denotes the foreground mask that is down-sampled to the corresponding size, $\circ$ represents the element-wise product, and Flatten($\cdot$) means the conversion from 2D feature map to 1D feature sequence.
To search for the relevant background styles, we enhance the foreground features through a transformer layer, where  \(\bm{F}_{c, f}^i\) serve as queries and \(\bm{F}_{c, b}^i\) serve as keys/values.
The stylized composite foreground features~\(\bm{\hat{F}}_{c, f}^i\) can be represented by
\begin{equation}
    \bm{\hat{F}}_{c, f}^i = \text{Transformer}(\bm{F}_{c, f}^i, \bm{F}_{c, b}^i, \bm{F}_{c, b}^i),
\end{equation}
where Transformer is a transformer encoder layer~\cite{vaswani2017attention}. 

Similar to \(\bm{{F}}_c^i\), the denoising feature map~\(\bm{F}_{z_t}^i\) can also be used to get the stylized denoising foreground features~$\bm{\hat{F}}_{z_t,f}^i$, so that the foreground features are stylized by relevant background styles.

\subsubsection{Stylized Feature Fusion} 
After extracting  $\bm{\hat{F}}_{c,f}^i$ and $\bm{\hat{F}}_{z_t,f}^i$ for \(i\in \{3,4\}\), we need to leverage both \(\bm{\hat{F}}_{c,f}^i\) and ~$\bm{\hat{F}}_{z_t,f}^i$ from dual encoders to help the denoising process. 
In particular, we first concatenate $\bm{\hat{F}}_{c,f}^i$ and $\bm{\hat{F}}_{z_t,f}^i$, and then pass them through a fully-connected layer to acquire the stylized foreground features \(\bm{\hat{F}}_{z_t, c}^i\).
The stylized foreground features \(\bm{\hat{F}}_{z_t, c}^i\) are then combined with the background of composite feature map \(\bm{F}_{c}^i\) and the denoising feature map $\bm{F}_{z_t}^i$ to guide the denoising steps.
The above steps could be formulated as 
\begin{align}
    \bm{\hat{F}}_{z_t, c}^i &= \text{FC}(\bm{\hat{F}}_{c,f}^i \oplus \bm{\hat{F}}_{z_t,f}^i), \\
    \hat{\bm{F}}_{z_t}^i &= \bm{F}_{z_t}^i + \text{Fold}(\bm{\hat{F}}_{z_t, c}^i) + \bm{{F}}_{c}^i \circ (1-\hat{\bm{M}}),
\end{align}
where FC($\cdot$) means the fully-connected layer, $\oplus$ means the concatenation between two vectors, Fold($\cdot$) means folding the \(\bm{\hat
{F}}_{z_t, c}^i\) into 2D foreground map.

\subsection{Objective Function}
First, we employ the standard noise loss from diffusion models~\cite{rombach2022high}, which aims to reconstruct the image feature within the latent space.
However, merely using noise loss can only reconstruct the composite image without changing the foreground style.
Therefore, we employ a combination of noise loss, AdaIN loss, and contrastive style loss, with the goal of attaining a balance between reasonable style and preservation of image structures/details. 
Besides, content loss is employed to assist in balancing noise loss and style losses.
In the following, we will detail four losses one by one.

\subsubsection{Noise Loss}
The denoising step of DDPM~\cite{ho2020denoising} is to remove noise from \(z'_{T}\) step by step and finally reconstruct the original composite input \(z'_{0}\). 
Therefore, the goal of the noise loss is to predict the noise in step $t$, which can be formulated as
\begin{equation} \label{eqn:noise_loss}
    \mathcal{L}_{LDM}:=\mathbb{E}_{z'_0, y, \epsilon \sim \mathcal{N}(0,1), t}\left[\left\|\epsilon-\epsilon_{\theta_1}\left(z'_{t}, t, \tau_{\hat{\theta}_2}(y)\right)\right\|_{2}^{2}\right],
\end{equation}
in which \(t\) is sampled from \(\{1,...,T\}\), $y$ is the condition information (\emph{i.e.}, composite image and foreground mask) in our problem. $\epsilon_{\theta_1}$ includes the model parameters of denoising U-Net, while $\tau_{\hat{\theta}_2}$ includes the model parameters of adaptive encoder and dual encoder fusion module.

\subsubsection{AdaIN Loss}
\label{sec:AdaIN}
Note that the noise loss in Eqn.~(\ref{eqn:noise_loss}) is calculated in the latent space, whereas the style losses cannot be directly calculated in the latent space.
Thus, we calculate two style losses based on the decoded image through decoder \(\mathcal{D}\), giving \(\bm{\hat{I}}_{0}=\mathcal{D}(\hat{z}_{0}^t)\).  \(\hat{z}_{0}^t=\left(z'_{t}-\sqrt{1-\bar{\alpha}_{t}} \epsilon_{\theta_1}\left( z'_{t},t,\tau_{\hat{\theta}_2}(y)\right)\right) / \sqrt{\bar{\alpha}_{t}}\)~\cite{ho2020denoising}, in which $\bar{\alpha}_{t}=\prod\limits_{s=1}^t {\alpha}_{s}$, $ {\alpha}_{s}=1-{\beta}_{s}$ and ${\beta}_{s}$ represents forward process variances. Based on $\bm{\hat{I}}_{0}$, we can easily calculate style losses. 

We utilize the AdaIN style loss~\cite{gatys2016image} to achieve consistency in multi-scale feature statistics (\emph{i.e.}, mean, standard deviation) between the background painting and the foreground in the harmonized image \(\hat{\bm{I}}_{0}\). 
The style loss can be written as
\begin{equation} \label{eqn:adain_loss}
    \begin{aligned} 
        \mathcal{L}_{AdaIN}= & \sum_{l=1}^{L}\left\|\mu\left(\phi^{l}\left(\hat{\bm{I}}_{0}\right) \circ \bar{\bm{M}}^{l}\right)-\mu\left(\phi^{l}\left(\bm{I}_{b}\right)\right)\right\|_2^{2}+ \\ & \sum_{l=1}^{L}\left\|\sigma\left(\phi^{l}\left(\hat{\bm{I}}_{0}\right) \circ \bar{\bm{M}}^{l}\right)-\sigma\left(\phi^{l}\left(\bm{I}_{b}\right)\right)\right\|_2^{2},
    \end{aligned}
\end{equation}
where \(\phi^{l}, l \in \{1,2,3,4\}\) represents the \(l\)-th ReLU\(\_ l \_ 1\) layer in a pre-trained VGG-19~\cite{simonyan2014very} network. 
\(\bm{I}_b\) is the complete background painting and \(\bar{\bm{M}}^{l}\) denotes the foreground mask that is down-sampled to the corresponding size. $\mu(\cdot)$ means the calculation of mean value and $\sigma(\cdot)$ means the calculation of standard deviation.

\begin{figure}[t]
  \centering
  \includegraphics[width=0.90\linewidth]{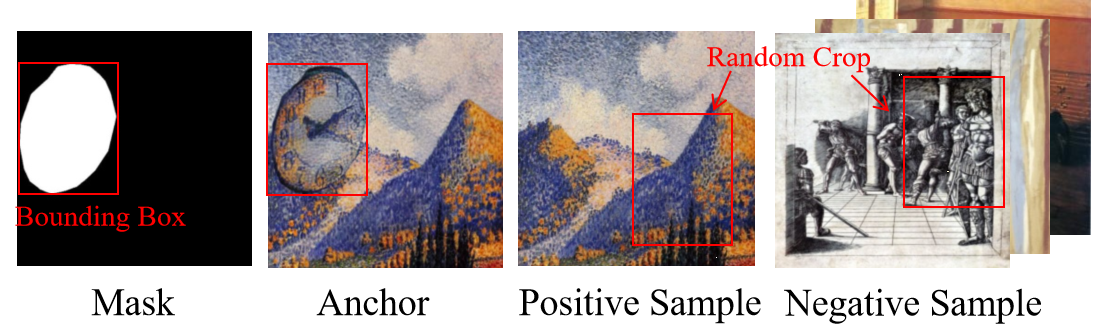}
  \caption{Construction of a triplet of anchor, positive sample, and negative sample for contrastive style loss.}
  \label{fig:contra}
\end{figure}

\subsubsection{Contrastive Style Loss}
To better migrate background style to foreground, we introduce another contrastive style loss, which is complementary with AdaIn loss. 
Contrastive style loss was first introduced into style transfer task by ~\cite{chen2021artistic},
which distinguishes the image rendered by the reference style from other styles.
Here, we adapt contrastive style loss to our painterly image harmonization task. 
Specifically, we construct a triplet of three elements: anchor, positive sample, and negative sample, as shown in \Cref{fig:contra}.
To acquire anchor, we first feed the harmonized output \(\hat{\bm{I}}_{0}\) into pre-trained VGG-19 network and extract the output feature map of ReLU\(\_ 3 \_ 1\) layer. 
Then, we crop the feature map with the downsampled mask followed by average pooling to obtain the foreground feature, which is projected to the anchor vector \(\bm{f}_{q}\). 
Through the same procedure, positive sample \(\bm{f}_{b}^{+}\) is extracted from the background painting \(\bm{I}_{b}\), so that \(\bm{f}_{q}\) and \(\bm{f}_{b}^{+}\) share the same style. 
And negative samples \(\bm{f}_{b}^{-}\) are extracted from other style images.
Given the triplet, we tend to pull close the anchor and positive example while separating the anchor from negative examples, which can be represented as
\begin{equation}\mathcal{L}_{CL}=-\log \left(\frac{\exp \left(\left(\bm{f}_{q}\right)^{T}  \left(\bm{f}_{b}^{+}\right) / \eta\right)}{\exp \left(\left(\bm{f}_{q}\right)^{T}   \left(\bm{f}_{b}^{+}\right) / \eta \right)+\sum_{\bm{f}_{b}^{-}} \exp \left(\left(\bm{f}_{q} \right)^{T} \left(\bm{f}_{b}^{-}\right) / \eta\right)} \right), \end{equation}\\
where the temperature \(\eta\) regulates the push and pull forces. We set \(\eta\) as 0.2 following~\cite{chen2021artistic}. 

\subsubsection{Content Loss}
In addition, when balancing between noise loss and style losses, chances are that the content details are excessively preserved, leading to insufficient style transfer. Therefore, we reduce the weight of noise loss and incorporate content loss~\cite{gatys2016image}, which is commonly used in style transfer tasks. The content loss can help preserve the high-level content information without the sacrifice of styles.
The content loss can be written as
\begin{equation}
\mathcal{L}_{con}=  \left\|\phi^{4}\left(\hat{\bm{I}}_{0}\right) -\phi^{4}\left(\bm{I}_{c}\right)\right\|_2^{2},
\end{equation}
where $\phi^{4}$ has been defined below Eqn.~(\ref{eqn:adain_loss}).

\subsubsection{Total Loss} By summarizing the noise loss, two style losses, and content loss, the total loss can be written as
\begin{equation}
    \mathcal{L}_{total}=\lambda_{1} \mathcal{L}_{LDM} +\mathcal{L}_{AdaIN}+\lambda_{2} \mathcal{L}_{CL}+ \mathcal{L}_{con},
\end{equation}
in which $\lambda_1$ and $\lambda_2$ are hyper-parameters. We empirically set them as 60 and 5  respectively.

\section{Experiments}
\subsection{Experiment Settings}
We train the adaptive encoder and the fusion module for 10 epochs with a batch size of 2.
We utilize Adam as the optimizer with the learning rate of \(2 \times 10^{-4}\).
During training, we resize the input images and the mask to \(512 \times 512\) and use the pretrained Stable Diffusion model~\cite{cong2022high} with the version of sd-v1-4.
We utilize the training data of COCO~\cite{lin2014microsoft} and WikiArt~\cite{nicholwikiart}. 
For more implementation details, please refer to the supplementary.

\subsection{Baselines}
Based on the target task, existing baselines can be categorized into three groups: painterly image harmonization~\cite{zhang2020deep, luan2018deep, cao2022painterly}, cross-domain composition methods~\cite{hachnochi2023cross, meng2021sdedit}, and artistic style transfer methods~\cite{huang2017arbitrary, liu2021adaattn, park2019arbitrary, deng2022stytr2, zhang2022inversion}.
The painterly image harmonization methods include DIB~\cite{zhang2020deep}, DPH~\cite{luan2018deep}, and PHDNet~\cite{cao2022painterly}.
The cross-domain composition methods include CDC~\cite{hachnochi2023cross} and SDEdit~\cite{meng2021sdedit}.
The artistic style transfer methods include AdaIN~\cite{huang2017arbitrary}, AdaAttN~\cite{liu2021adaattn}, SANet~\cite{park2019arbitrary}, StyTr2~\cite{deng2022stytr2}, and InST~\cite{zhang2022inversion}.
Among them, CDC~\cite{hachnochi2023cross}, SDEdit~\cite{meng2021sdedit}, and InST~\cite{zhang2022inversion} are diffusion-based methods.

For the first and the second groups, these works can stylize a certain region, so we directly compare them with our results.
However, for the third group, these works stylize the entire photographic image. 
To adapt artistic style transfer methods to our task, we stylize the content image according to the background image, followed by cutting and pasting the stylized foreground object onto the background image. 
We set \(Strength\) to 0.7 by default to control total inference steps for our model. More details of implementations including hyper-parameters of baselines are in the supplementary.

\begin{figure*}[t]
  \centering
  \includegraphics[width=0.83\linewidth]{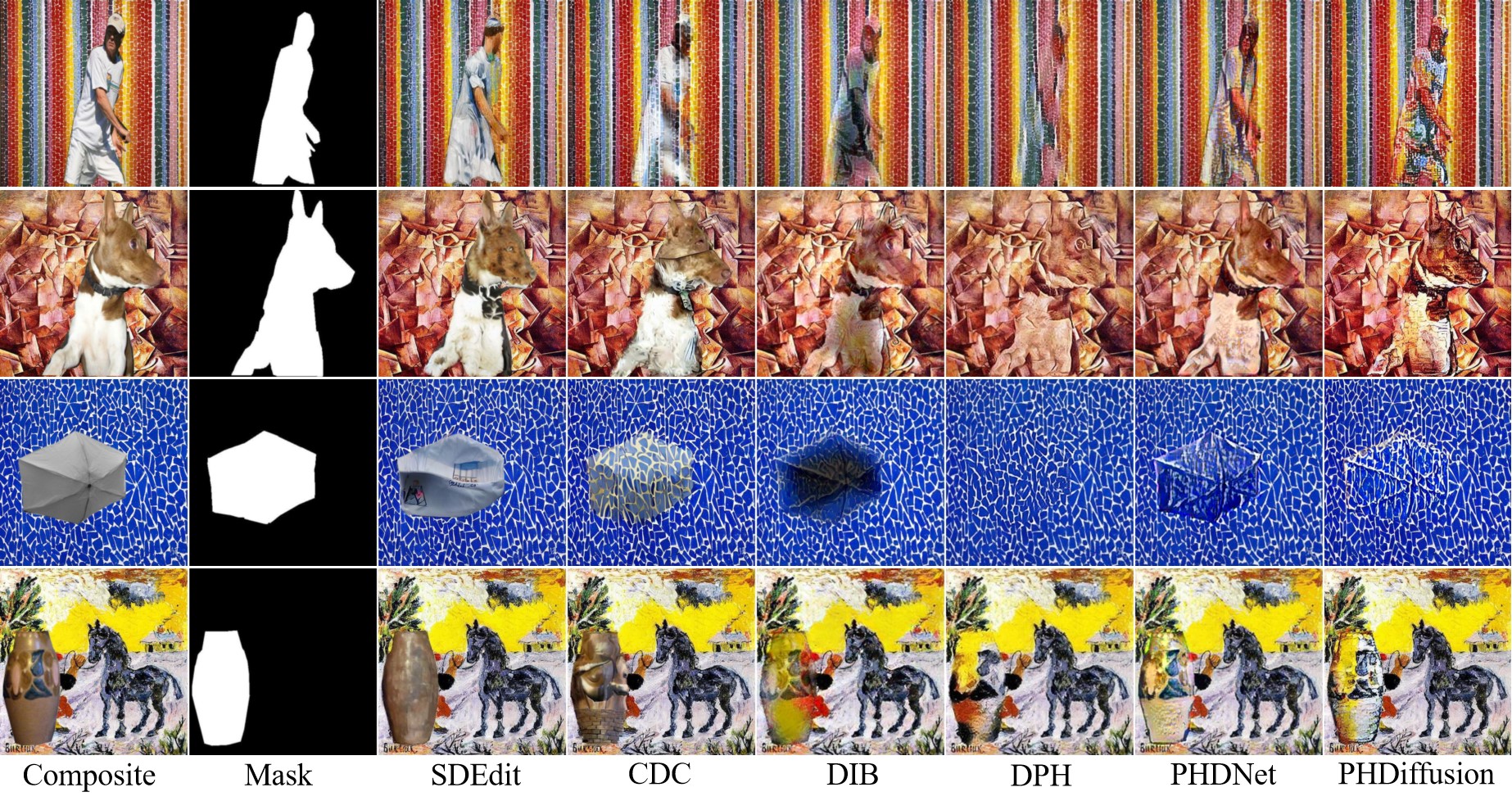}
  \caption{From left to right, we show the composite image, mask, harmonized results of SDEdit, CDC, DIB, DPH, PHDNet and our PHDiffusion. Best viewed in color and zoom in.}
  \label{fig:painterly}
\end{figure*}

\begin{figure*}[t]
  \centering

  \includegraphics[width=0.83\linewidth]{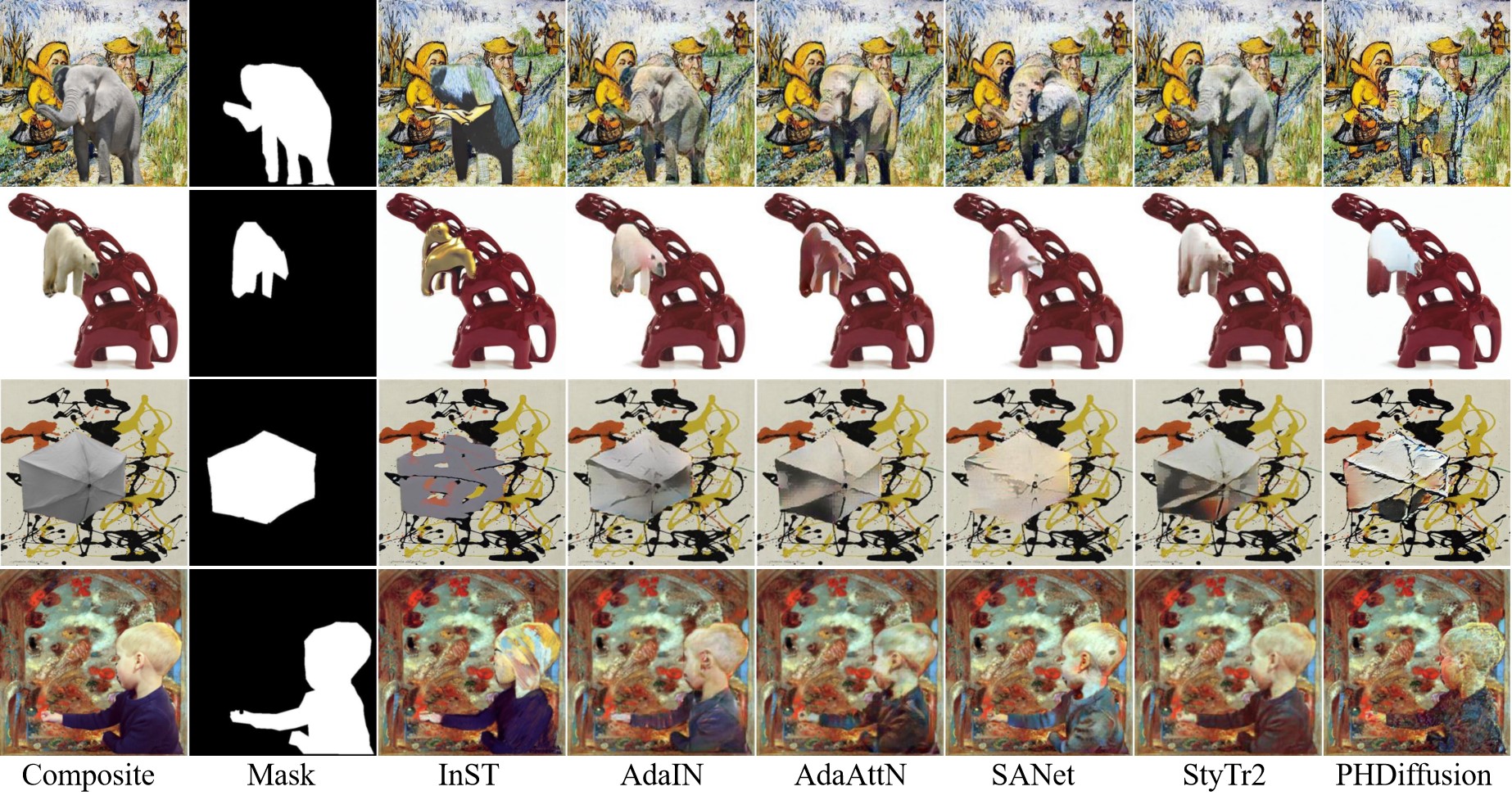}
  \caption{From left to right, we show the composite image, mask, example results for InST, AdaIN, AdaAttN, SANet, StyTr2 and our PHDiffusion. Best viewed in color and zoom in.}
  \label{fig:transfer}
\end{figure*}

\subsection{Comparisons with Baselines}

\begin{table*}[t]
\centering
{
\begin{tabular}{c|ccccc|ccc|cc|c}
\hline
        & AdaIN  & AdaATTN & SANet  & StyTr2 & InST   & DIB    & DPH    & PHDNet & SDEdit & CDC    & PHDiffusion   \\ \hline
BT      & -0.224 & 0.180   & 0.596  & 0.912  & -1.779 & -1.123 & 1.376  & 1.811  & -2.476 & -1.863 & 2.590  \\\hline
\end{tabular}}
\caption{ Comparisons with baselines. "BT" stands for B-T score.}
\label{tab:baseline}
\end{table*}

\subsubsection{Visualization Analysis} 
Compared with the first group of baselines, painterly harmonization methods, we can refer to \Cref{fig:painterly} for the visualization results. It is shown that our PHDiffusion can endow the foreground with more abundant and coherent styles (row 1, 2, 3, 4). For example, as illustrated in row 1, our PHDiffusion can not only learn the local dotted textures, but also learn the global stripe arrangement.
Besides, our method can strike great balance between content and style. 
For content preservation, our method holds more semantic edge information (row 1, 2, 3), and also maintains more refined details (row 1). 
In row 3, our method preserves the clear umbrella frame while capturing the textures closest to the background.
However, DPH loses details and semantic edge, while DIB and PHDNet fail to transfer coherent textures.

Compared with the second group, cross-domain composition methods, as illustrated in \Cref{fig:painterly} (left), we can see that SDEdit and CDC struggle to transfer sufficient style while keeping original content (row 1, 3, 4). In row 1, the content has already changed sharply while the style fails to be adapted to the target. Though CDC has learnt some dotted textures, the style and content exhibit a lack of cohesion, resembling distinct layers. And in row 3 and row 4, the content is lost to some extent with details unrecognized. The poor performance of baselines is probably caused by the stochastic nature of diffusion model, which can be a double-edged sword for the tasks that require handling delicate details, since it is very hard to adjust hyper-parameters, (\emph{e.g.}, strength) to balance between style and content without proper guidance. In contrast, our method provides more effective guidance for the denoising process.

Compared with the third group, artistic transfer methods, as shown in \Cref{fig:transfer}, it can be seen that InST loses content and exhibits style incompatible with the background (row 1, 4). Other style transfer methods can not produce adequate styles. 
Our method not only produces textures that highly match the background (row 1, 2, 3, 4), but also enables the overall color distribution to be strongly correlated with the background (row 1, 3), leading to better visual harmony.

The great performance of our PHDiffusion is attributed to two aspects. 
Firstly, for producing sufficient styles, our DEF module can query backgrounds for reasonable styles and utilize prior knowledge in pretrained stable diffusion model. 
Secondly, for balancing content and style, the combination of noise loss, content loss, and style losses enable the adaptive encoder and DEF module to store appropriate guiding information.

\begin{table}
\centering
\setlength\tabcolsep{9pt}
\resizebox{0.7\columnwidth}{!}{
\begin{tabular}{c|l|c}
\hline
Version  & Method      & BT \\ \hline
V1    & w/o  \(\mathcal{L}_{CL}\)&  0.636  \\
V2    & w/o  \(\mathcal{L}_{AdaIN}\)   & -1.584   \\
V3    & w/o TA,TU     & -0.421   \\
V4    & w/o TA        &  0.182  \\
V5    & w/o TU        &  -0.112  \\
V6    & full           &   1.300 \\ \hline
\end{tabular}}
\caption{The results of the ablation experiments. "TA" and "TU" stand for the transformer layer in adaptive encoder and U-Net encoder respectively. "BT" stands for B-T score.}
\label{tab:ablation}
\end{table}

\begin{figure}[t]
  \centering
  \includegraphics[width=\linewidth]{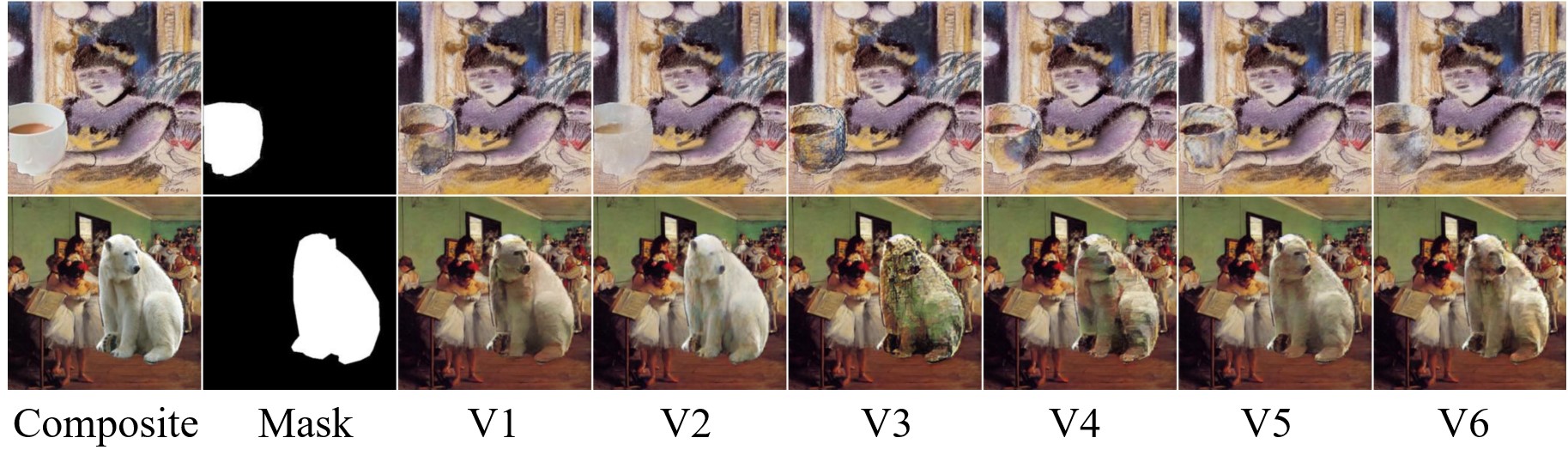}
  \caption{Examples of ablation experiments. Best viewed in color and zoom in.}
  \label{fig:ablation}
\end{figure}

\subsubsection{User Study}
We also conduct a user study to compare the effectiveness of various methods, following~\cite{cao2022painterly}.
Specifically, we randomly select 100 content images from COCO~\cite{lin2014microsoft} and 100 style images from WikiArt~\cite{nicholwikiart} to generate 100 composite images.
Given each composite image, we can obtain 11 harmonized results including 10 baselines and our method. Then pairwise comparisons are conducted, resulting in 5,500 image pairs. 
We invite 50 users to identify the more harmonious one in each pair. 
Finally 275,000 comparison results are collected, followed by using the Bradley-Terry (B-T) model~\cite{bradley1952rank,lai2016comparative} to calculate an overall ranking of all methods. As presented in \Cref{tab:baseline}, our PHDiffusion achieves the highest B-T score.

\subsection{Ablation Studies}

As described in \Cref{sec:method}, our PHDiffusion exploits an adaptive encoder along with the dual encoder fusion module to guide the denoising process and two style losses to balance the content and style. 
Therefore, in this section, we demonstrate their effectiveness, and report B-T score in \Cref{tab:ablation} and visual results in \Cref{fig:ablation}. 
For the effectiveness of style losses, we conduct experiments without contrastive style loss (V1) or AdaIN loss (V2). 
For the effectiveness of transformer layer in DEF module, we conduct experiments in the following three settings:
(1) remove transformer layers in both encoders (V3);
(2) remove transformer layer in adaptive encoder (V4);
(3) remove transformer layer in U-Net encoder (V5).

Comparing V1, V2, and V6 in \Cref{tab:ablation}, we find that the AdaIN loss is more important for style transfer, while the contrastive style loss assists in capturing more reasonable styles. Moreover, by comparing V3, V4, V5, and V6 in \Cref{tab:ablation}, we can find that the transformer layers in U-Net encoder and adaptive encoder are both helpful to generate reasonable styles, and exploiting transformer layer in both encoders can further boost the painterly image harmonization.

For visual results in \Cref{fig:ablation}, comparing V1, V2, and V6, it is observed that AdaIN loss (V1) can learn styles that appear to be blended, while contrastive style loss (V2) tends to learn fine textures (more haziness for cup and more textures of the fur for bear) while maintaining the original color. So the combination of two style losses helps the transformer capture adequate local textures and fine global styles.
Comparing V3 and V4 (V3 and V5) in \Cref{fig:ablation}, we can find that the transformer can help learn more consistent and reasonable styles from the background.
Comparing V4, V5, and V6 in \Cref{fig:ablation}, it is observed that the adaptive encoder prefers more subdued color (V5) while the U-Net encoder tends to perform more exaggerated color (V4). Balancing them can achieve more reasonable and harmonized styles for our final result (V6).

\section{Conclusion}
In this work, we have introduced diffusion model into painterly image harmonization. 
We have proposed a novel Painterly Harmonization stable Diffusion model (PHDiffusion), in which the denoising process of diffusion is under the guidance of lightweight adaptive encoder and dual encoder fusion. 
Experiments have demonstrated that our approach can simultaneously preserve detailed content and produce sufficient styles, surpassing the state-of-the-art methods.

\begin{acks}
 The work was supported by the Shanghai Municipal Science and Technology Major / Key Project, China (Grant No. 20511100300 / 2021SHZDZX0102) and the National Natural Science Foundation of China (Grant No. 62076162).
\end{acks}

\bibliographystyle{ACM-Reference-Format}

\end{document}


\title{Supplementary Material for Painterly Image Harmonization using Diffusion Model}

\author{Lingxiao Lu}
\affiliation{%
    \institution{MoE Key Lab of Artificial Intelligence, Shanghai Jiao Tong University}
    \country{China}
}
\email{lulingxiao@sjtu.edu.cn}

\author{Jiangtong Li}
\affiliation{%
    \institution{MoE Key Lab of Artificial Intelligence, Shanghai Jiao Tong University}
    \country{China}
}
\email{keep_moving-lee@sjtu.edu.cn}

\author{Junyan Cao}
\affiliation{%
    \institution{MoE Key Lab of Artificial Intelligence, Shanghai Jiao Tong University}
    \country{China}
}
\email{joy_c1@sjtu.edu.cn}

\author{Li Niu}
\authornote{Corresponding authors}
\affiliation{%
    \institution{MoE Key Lab of Artificial Intelligence, Shanghai Jiao Tong University}
    \country{China}
}
\email{ustcnewly@sjtu.edu.cn}

\author{Liqing Zhang}
\authornotemark[1]
\affiliation{%
    \institution{MoE Key Lab of Artificial Intelligence, Shanghai Jiao Tong University}
    \country{China}
}
\email{zhang-lq@cs.sjtu.edu.cn}

\maketitle

\appendix

\begin{figure*}[ht]
  \centering
  \includegraphics[width=0.93\linewidth]{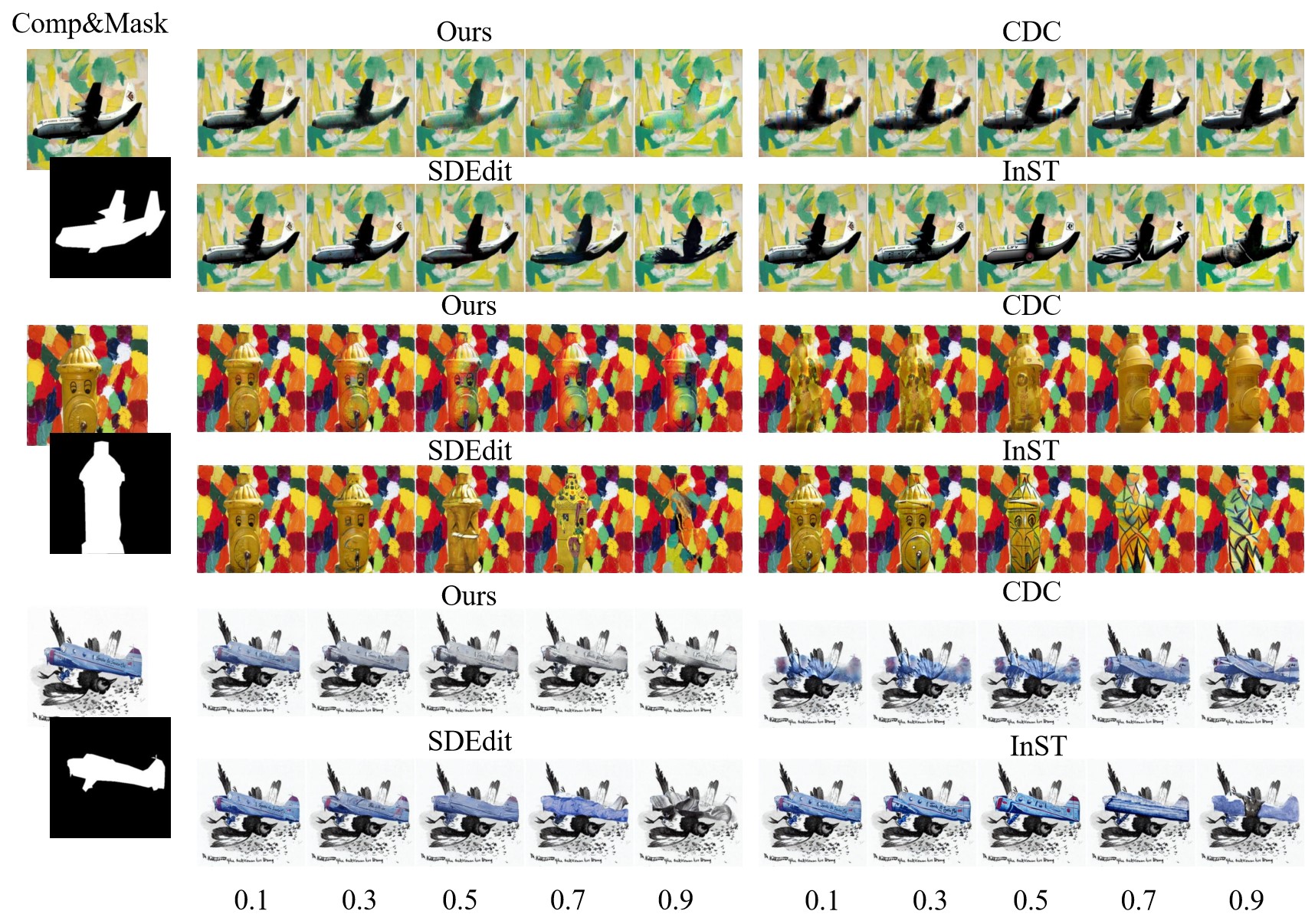}
  \caption{The results of adjusting \(\bm{Strength}\) for controlling the degree of style transfer. \(\bm{Strength}\) is set to \(\bm{0.1,0.3,0.5,0.7,0.9}\) from left to right for each method. We also present the results of Ours, CDC, SDEdit, and InST for comparison.}
  \label{fig:strength}
\end{figure*}

In the supplementary, we will first introduce the dataset and implementation details in \Cref{sec:imple}. 
Then the hyper-parameter \(Strength\) will be studied for style strength control in \Cref{sec:stre_ctrl}. The visualization of attention maps in the transformers of our dual encoder fusion module will be explained in \Cref{sec:atten}.
We will also provide more details of implementing baselines and offer more visual comparison results in \Cref{sec:comp_base}. 
Finally, we will discuss the limitations of our method in \Cref{sec:limit}.

\section{Dataset and Implementation Details}\label{sec:imple}

We conduct experiments on two benchmark datasets, \emph{i.e.}, COCO~\cite{lin2014microsoft} and WikiArt~\cite{nicholwikiart}, where COCO is a large-scale photograph dataset with the instance segmentation annotation for 80 different object categories and WikiArt is a large-scale digital art dataset consisting of 27 distinct styles.
These two datasets are used to produce composite images by inserting photographic foreground objects from COCO into painterly backgrounds from WikiArt.

In detail, to obtain the foreground object with proper size and resolution, we select 9,100 foreground images from the COCO dataset, whose foreground ratio is between 0.05 and 0.3, and width and height are $\geq$ 480.
Moreover, we select 37,931 background images from the WikiArt dataset, whose width and height are $\geq$ 512.
During training, we use instance annotation to extract the foreground objects from the foreground images and then place it onto a randomly chosen painterly background from the background images, leading to 37,931 composite images in each epoch. 
Finally, all the composite image are resized to $512 \times 512$ for training.
This process can produce composite images with discordant visual elements.

Our network is implemented using Pytorch 1.11.0. 
And the training process is executed on an Ubuntu 20.04 LTS operating system, utilizing a computing environment comprising of 32GB memory, Intel Xeon Silver 4116 CPU, and two GeForce RTX 3090 GPUs.

\section{Strength Control}\label{sec:stre_ctrl}
\(Strength\) is a hyper-parameter that decides the total step in the inference process. For example, the total step is equal to 35 when \(Strength\) is 0.7 and default total step is 50. The larger total step means more noise to be added and removed, leading to larger variability, so that the guidance from condition information could have greater impact on the harmonized results. 

We observe that when \(Strength\) grows larger in a proper range (\emph{i.e.}, 0.1 - 0.7) , the style of our harmonized result gets transferred gradually while the content details are barely changed.
However, as the \(Strength\) changes, other diffusion model baselines cannot balance the style and content.
In \Cref{fig:strength}, we visualize how the harmonization results changes as the strength changes.
In detail, we compare the harmonization results of SDEdit~\cite{meng2021sdedit}, CDC~\cite{hachnochi2023cross}, InST~\cite{zhang2022inversion}, and our PHDiffusion with the \(Strength\) ranging from \(0.1\) to \(0.9\).
Recall that the inference process of diffusion model is under control of the \(Strength\), where the smaller the \(Strength\) is, the smaller the denoising step is.
If the denoising process is guided improperly or without guidance, the style and the content cannot be balanced as the \(Strength\) changes.
In detail, for SDEdit~\cite{meng2021sdedit} and InST~\cite{zhang2022inversion} in \Cref{fig:strength}, as the \(Strength\) becomes larger, the style is more sufficient while the content is destroyed. 
For CDC~\cite{hachnochi2023cross}, since the denoising process is guided by the composite image, as the \(Strength\) gets larger, the content details are more preserved while the style is ignored; however, as the \(Strength\) gets smaller, the style is more sufficient while the content is destroyed.
Besides, the balanced point for CDC~\cite{hachnochi2023cross} is also quite vulnerable.

However, if this process is guided by our DEF module, it can be tailored to painterly image harmonization with smooth transition between original and target styles without destroying content. 
Specifically, from our harmonization results in \Cref{fig:strength}, we observe that the styles are progressively strengthened while content is well-preserved (details such as eyes in the second example and words on the airplane in the third example remain visually clear). 
This proves that diffusion process is highly controllable and our mechanism can provide powerful guidance for the diffusion process.

\begin{figure}[t]
  \centering
  \includegraphics[width=\linewidth]{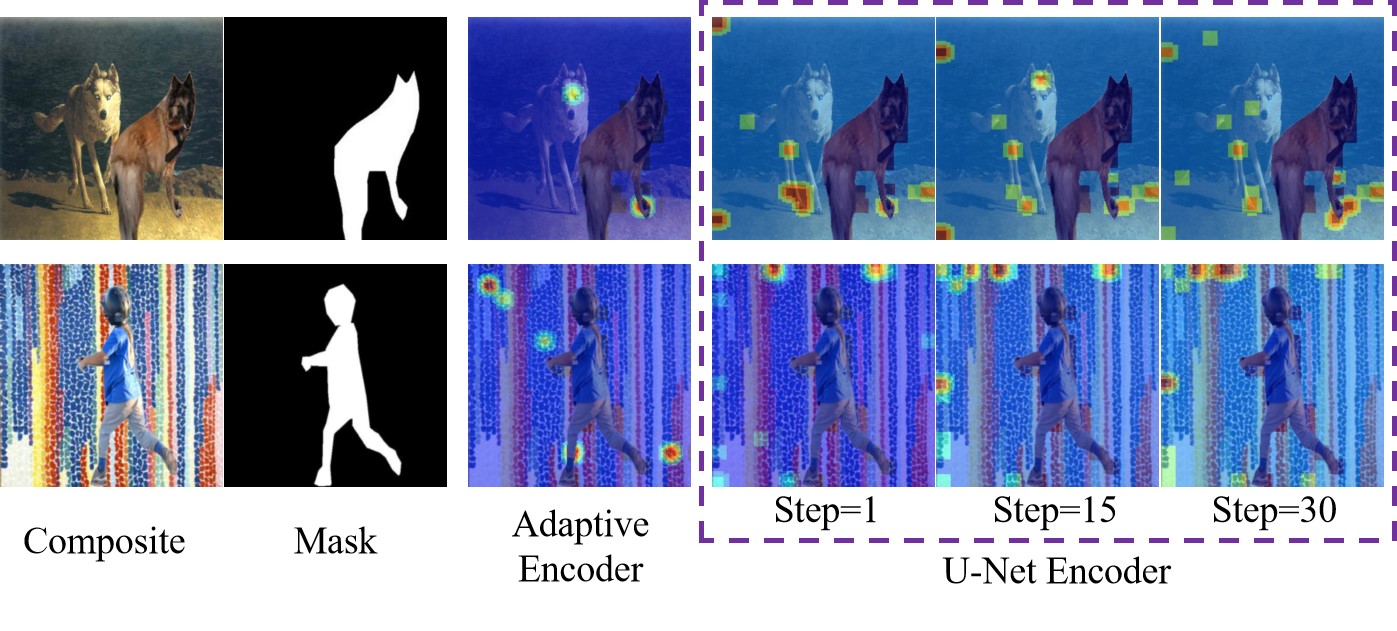}
  \caption{The attention maps in DEF module for Adaptive Encoder and U-Net Encoder. For U-Net Encoder, we present its attention maps in different timesteps.}
  \label{fig:attn}
\end{figure}

\section{Visualization of Attention Maps for Dual Encoder Fusion}\label{sec:atten}
To better understand the transformers in DEF module, we visualize the attention maps of transformers during inference in \Cref{fig:attn}. 
Since the feature maps in adaptive encoder remain the same during  multi-step inference, so its attention maps also remain the same. 
In contrast,  the feature maps in U-Net encoder are updated  during  multi-step inference, so the attention maps vary in different steps. 
Therefore, in~\Cref{fig:attn}, we show how the attention maps in U-Net encoder changes during denoising steps. 
Specifically, each attention map is obtained by averaging the attention maps from all attention heads, and then resized to the original resolution.
In~\Cref{fig:attn}, it can be observed that, for composite image whose background has similar objects with foreground, these objects can be detected and attended by the transformer layer, so that these objects in the background usually have larger weights. 
For example, in row 1 of~\Cref{fig:attn}, the wolf in the background, which is similar to the foreground dog, gains attention in both adaptive encoder and U-Net encoder. 
For U-Net encoder in different steps, the wolves are all attended. 
Moreover, the ground surrounding the dog, which has similar color to the dog, is also attended by our DEF module. 
Besides, for composite image whose background has pure textures (row 2), the DEF module seems to pay attention to the background randomly for both adaptive and U-Net encoders to capture the overall pattern.
These visualization results again prove that our DEF module can focus on meaningful background regions and provide rational guidance during the denoising steps.

\section{Comparison with Baselines}\label{sec:comp_base}

\subsection{Details of Diffusion-based Baselines}

For all the diffusion-based baselines, since the \(Strength\) has great influence on the harmonization results, we adjust the \(Strength\) and select the optimal outcome for comparison. 
Therefore, we set \(Strength\) to 0.5, 0.5 and 0.7 for SDEdit~\cite{meng2021sdedit}, CDC~\cite{hachnochi2023cross} and InST~\cite{zhang2022inversion}, respectively, which are the best results for balancing content and style. 
For our PHDiffusion, we choose $Strength=0.7$ by default.

Besides, to adapt style transfer method InST~\cite{zhang2022inversion} for painterly image harmonization, for training process, we train the proposed small conditional network on WikiArt while freezing the diffusion model, after which the model is able to learn styles in WikiArt better.
Moreover, during inference, we first add noise for \(T\) steps to the composite image in forward process. 
During each backward step, we exploit the corresponding background in $i$-th forward step to replace the background of predicted images in ($T-i$)-th background step, which aims to preserve the background and only adapt the foreground to satisfy the demand of painterly image harmonization.

\begin{figure}[t]
  \centering
  \includegraphics[width=0.9\linewidth]{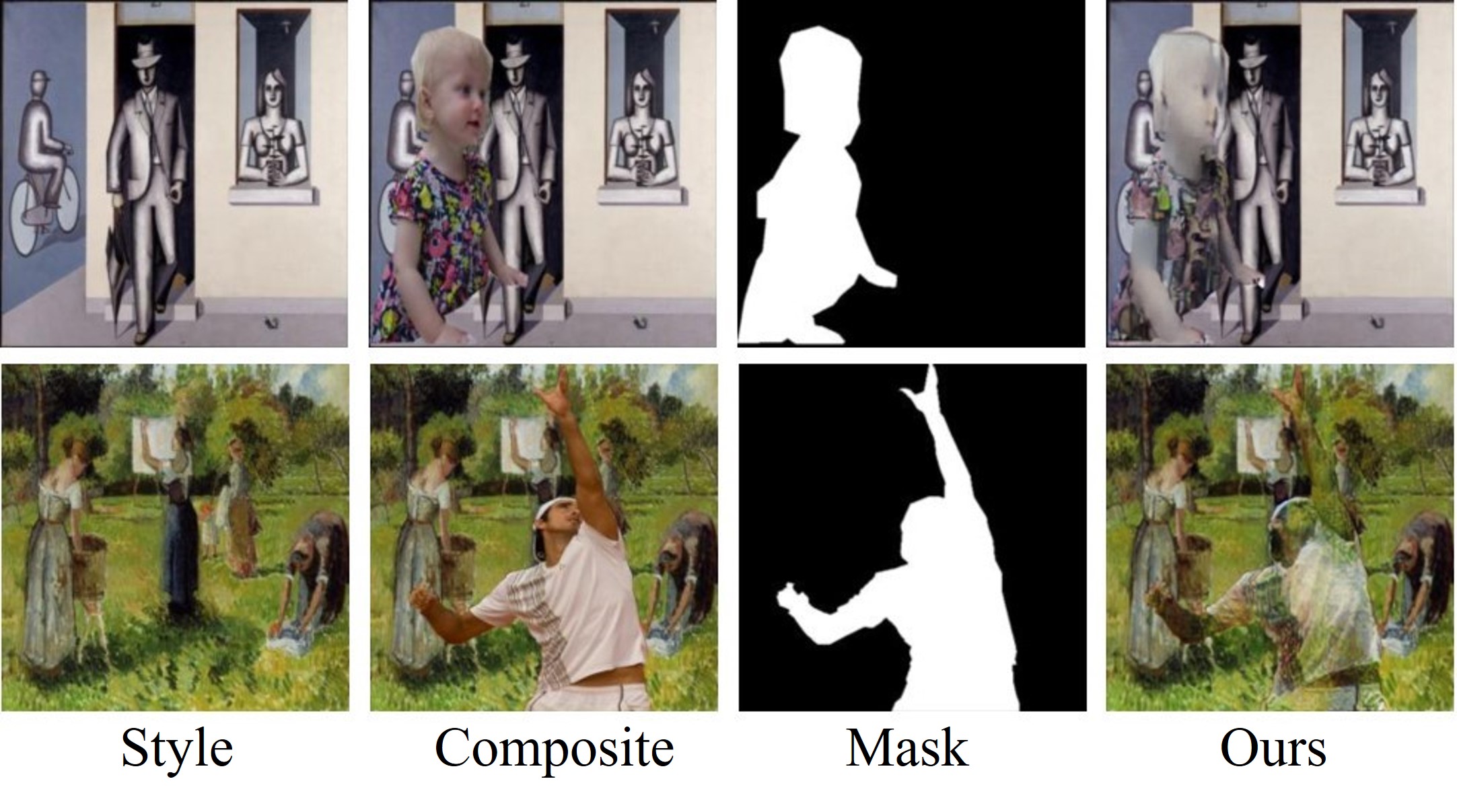}
  \caption{Example failure cases of our PHDiffusion.}
  \label{fig:limit}
\end{figure}

\subsection{More Visual Results}
\begin{figure*}[t]
  \centering
  \includegraphics[width=0.9\linewidth]{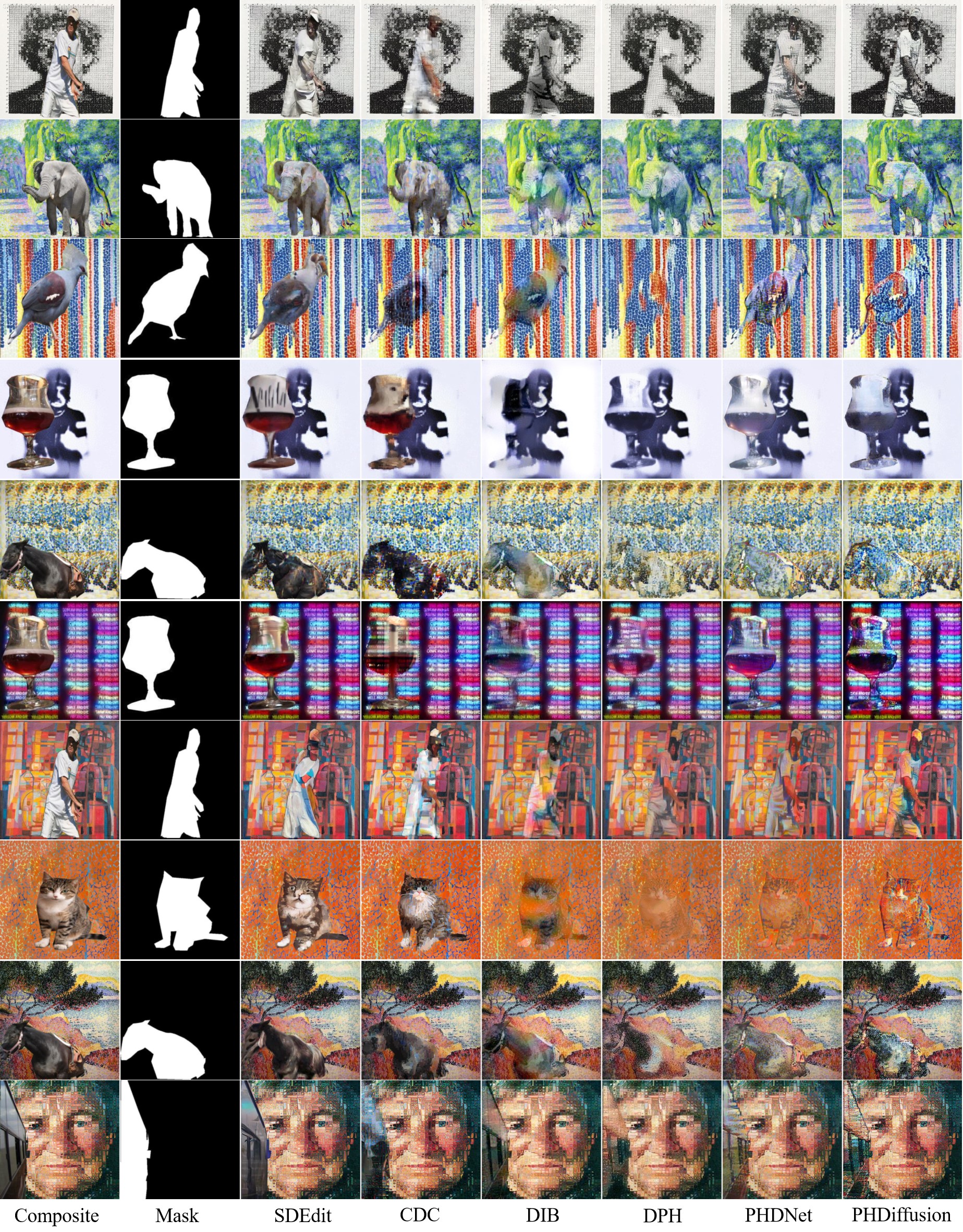}
  \caption{From left to right, we show the composite image, mask, harmonized results of SDEdit~\cite{meng2021sdedit}, CDC~\cite{hachnochi2023cross}, DIB~\cite{zhang2020deep}, DPH~\cite{luan2018deep}, PHDNet~\cite{cao2022painterly}, and our PHDiffusion. Best viewed in color and zoom in.}
  \label{fig:painter}
\end{figure*}

\begin{figure*}[t]
  \centering
  \includegraphics[width=0.9\linewidth]{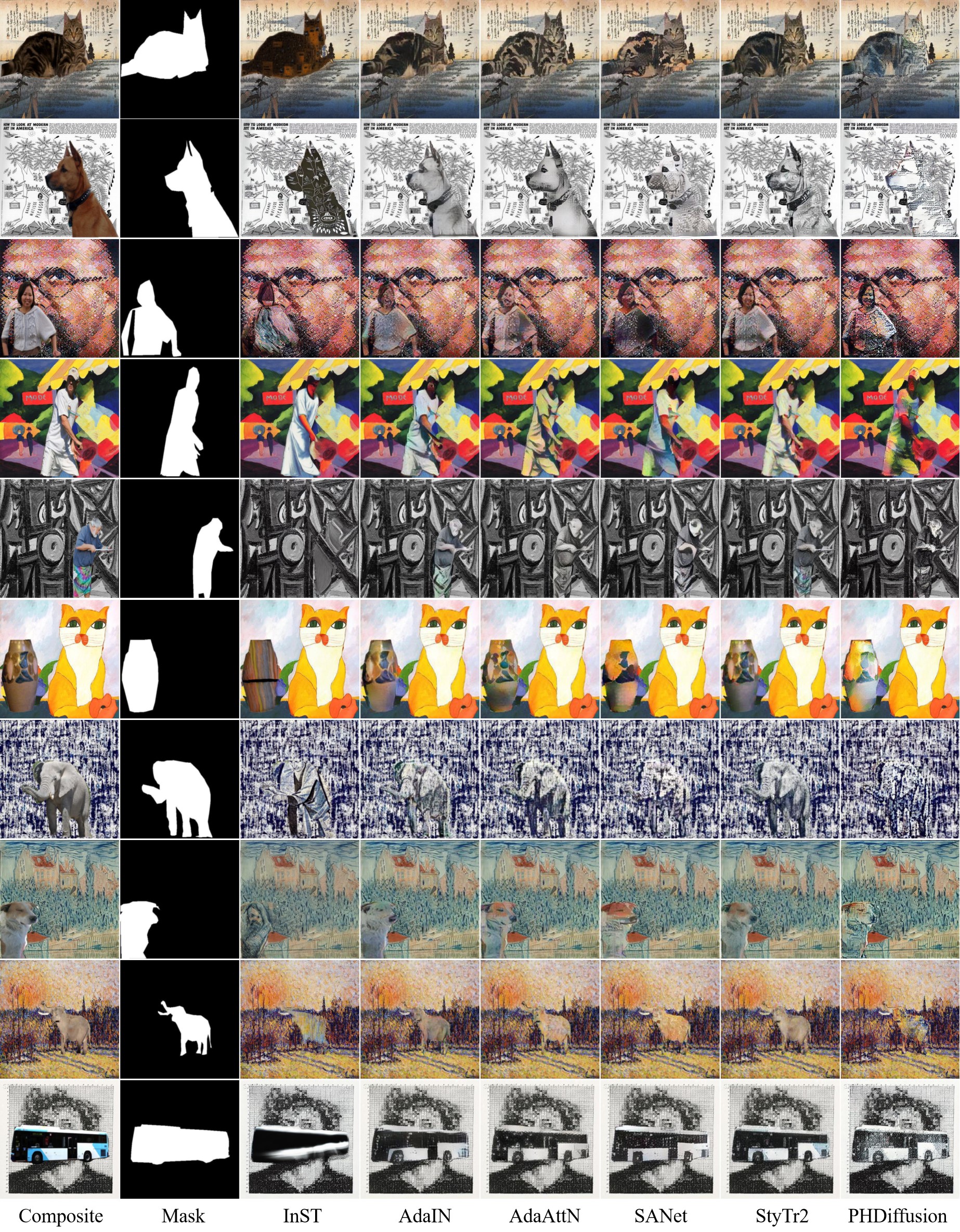}
  \caption{From left to right, we show the composite image, mask, example results for InST~\cite{zhang2022inversion}, AdaIN~\cite{huang2017arbitrary}, AdaAttN~\cite{liu2021adaattn}, SANet~\cite{park2019arbitrary}, StyTr2~\cite{deng2022stytr2}, and our PHDiffusion. Best viewed in color and zoom in.}
  \label{fig:transfer}
\end{figure*}

We provide more visual results to compare with other baselines.
As we have introduced in the main submission, we have three groups of baselines. 
The first group contains painterly image harmonization methods,  DIB~\cite{zhang2020deep}, DPH~\cite{luan2018deep}, and PHDNet~\cite{cao2022painterly}. 
The second group includes cross-domain composition methods,  CDC~\cite{hachnochi2023cross} and SDEdit~\cite{meng2021sdedit}. 
And the third group, artistic style transfer, consists of AdaIN~\cite{huang2017arbitrary}, AdaAttN~\cite{liu2021adaattn}, SANet~\cite{park2019arbitrary}, StyTr2~\cite{deng2022stytr2}, and InST~\cite{zhang2022inversion}. 
The results for the first and second group are shown in \Cref{fig:painter}, while the results for the third group are shown in \Cref{fig:transfer}.

In \Cref{fig:painter}, it can be seen that DIB~\cite{zhang2020deep}, DPH~\cite{luan2018deep} and PHDNet~\cite{cao2022painterly} can also achieve harmonization to some extent, but the learnt textures are not as accurate as ours (row 1, 3, 4, 5, 10). 
Besides, our PHDiffusion can capture more global styles, thus tending to be more naturally blended with the background (row 2, 6, 7, 8). 
Specifically, our PHDiffusion is able to maintain more semantic information and content details (\emph{e.g.}, the stripes on the body of the cat in row 8 and the pattern on the shirt of the man in row 7).
And for cross-domain methods in \Cref{fig:painter}, it is obvious that SDEdit~\cite{meng2021sdedit} tends to directly copy the content of foreground objects (row 2, 3, 6, 9, 10), and the style is not compatible with the background images. 
Besides, CDC~\cite{hachnochi2023cross} fails to preserve the content details (row 1, 3, 5, 9, 10).
The comparison between our PHDiffusion and other cross-domain methods (\emph{i.e.}, SDEdit~\cite{meng2021sdedit} and CDC~\cite{hachnochi2023cross}) turns out that our method outperforms them in both style and content.

For the third group in \Cref{fig:transfer}, it can be observed that InST~\cite{zhang2022inversion} has lost too much content details. 
Though its learnt style is quite consistent with the background, the harmonization results from InST~\cite{zhang2022inversion} are still less realistic (row 1, 4, 6).
Moreover, for AdaIN~\cite{huang2017arbitrary}, AdaAttN~\cite{liu2021adaattn}, SANet~\cite{park2019arbitrary}, and StyTr2~\cite{deng2022stytr2}, they can also partially migrate the background style, however, the styles in our results are obviously more harmonized (row 3 to row 9). 
For various types of backgrounds, our harmonization results always behave well in holding semantic information (row 3, 7, 9). 
Overall, for our harmonization results, the inserted foreground objects can be better integrated into the background, making the whole harmonized images appear to be intact artistic paintings.

\section{Limitations}\label{sec:limit}
Generally speaking, our method is capable of producing visually appealing and harmonious results, however, some types of foreground objects such as human faces are still hard to be in great harmony with the backgrounds. Since human faces have delicate details and we are very sensitive to the subtle changes in human faces,
it is very difficult to sufficiently stylize the human faces while preserving their delicate details. 

\bibliographystyle{ACM-Reference-Format}